\documentclass[a4paper, 10pt]{article}

\usepackage{amssymb}
\usepackage{graphicx}
\usepackage{amsmath}
\usepackage{bm}
\usepackage{algorithm}
\usepackage{algorithmic}
\usepackage{url}
\usepackage{threeparttable}

\begin{document}

\noindent {\Large \bf Estimation of the number of clusters on $d$-dimensional sphere}

\vspace{1cm}

\noindent {\bf Kazuhisa Fujita}$^{12}$

\noindent $^1$Komatsu University, Komatsu, Ishikawa, Japan

\noindent $^2$University of Electro-Communications, Chofu, Tokyo, Japan

\noindent Correspondence: Kazuhisa Fujita Address: Komatsu University, 10-10 Doihara-machi, Komatsu, Ishikawa, Japan. E-mail: kazu@spikingneuron.net

\section*{Abstract}

Spherical data is distributed on the sphere.
The data appears in various fields such as meteorology, biology, and natural language processing.
However, a method for analysis of spherical data does not develop enough yet.
One of the important issues is an estimation of the number of clusters in spherical data.
To address the issue, I propose a new method called the Spherical X-means (SX-means) that can estimate the number of clusters on $d$-dimensional sphere.
The SX-means is the model-based method assuming that the data is generated from a mixture of von Mises-Fisher distributions.
The present paper explains the proposed method and shows its performance of estimation of the number of clusters.

\section*{keywords}
clustering, k-means, spherical data, find k

\section{Introduction}
Spherical (directional / periodical) data is a set of data points on $d$-dimensional sphere.
In other words, spherical data is only represented by a direction.
Spherical data is found in many fields such as meteorology, biology, and natural language processing.
Wind direction at geographical location \cite{Carta:2008,Masseran:2013}, direction of animals' trajectories \cite{Calderara:2011,Domenici:2008}, a unit vector representing a text document \cite{Dhillon:2001}, and hue which is one coefficient of HSV and HSL color models are typical directional data.
In these fields, a clustering method for spherical data is required as an important analysis tool.
The spherical $k$-means (sk-means) \cite{Dhillon:2001} and the expectation maximization algorithm with a mixture of von Mises-Fisher distributions \cite{Banerjee:2003,Banerjee:2005} are famous clustering methods for spherical data.
These methods can partition spherical data but not estimate the number of clusters $k$.

Clustering is used in various application fields such as data mining, pattern recognition, computer vision, etc.
One of the typical methods of clustering is the $k$-means \cite{MacQueen:1967}.
The $k$-means is widely used because it is a simple and effective method.
However, the $k$-means cannot estimate the number of clusters.
In order to estimate the number of clusters, the X-means has been developed by Pelleg and Moore \cite{Pelleg:2000}.
The X-means is a useful and fast method for estimating $k$.
However, it is difficult to apply to data not sampled from a distribution with a Gaussian mixture because the X-means assumes data generated from a Gaussian mixture.

In the present study, I propose the novel method called the spherical X-means (SX-means) that can estimate the number of clusters and partition data on $d$-dimensional sphere.
The SX-means is inspired by the X-means.
In the SX-means, the data points are assumed to be generated from a mixture of von Mises-Fisher distributions.
This paper explains the algorithm of the SX-means and shows the results of estimated $k$ from synthetic data and real-world data by the SX-means.

\section{Related work} % (fold)
\label{sec:related_work}

The $k$-means is the most famous algorithm for clustering.
However, we have to decide the number of clusters $k$ in a dataset before clustering.
The determination of $k$ is one of the most challenging and difficult problems in cluster analysis.

Many algorithms to estimate $k$ have been proposed.
The elbow method is one of the commonly used methods to find $k$.
The elbow method manually determines $k$ according to the visualized curve of the criterion function of $k$ such as the sum of squared errors and the percentage of variance.
The elbow point, which is the point of maximum curvature, is regarded as the optimal $k$.
However, we cannot clearly determine $k$ from the curve when the curve is fairly smooth \cite{Shi:2021}.
Rousseeuw \cite{Rousseeuw:1987} has proposed the silhouette method to find $k$.
The silhouette method uses the silhouette coefficient that is a measure of how close one data point is to the others in the same cluster compared to the data points in its neighbor cluster, and determines $k$ from the plot of the silhouette coefficient and its average.
Kingrani et al. \cite{Kingrani:2017} have determined $k$ using the diversity measured by Rao's quadratic entropy.
The ISODATA proposed by Ball and Hall \cite{Ball:1965} is one of the methods to find $k$.
In the ISODATA, the number of data points in a cluster, the average distance of data points, and the standard deviation of a cluster dominate increase or decrease $k$.
Pelleg and Moore \cite{Pelleg:2000} have proposed the method to find $k$ called the X-means.
The X-means is based on the $k$-means and finds $k$ according to the BIC.
It performs faster than repeatedly using iterated k-means.
Hamerly and Elkan \cite{Hamerly:2003} have proposed the G-means developed based on the X-means and uses Anderson-Darling test to estimate $k$.
Feng and Hamerly \cite{Feng:2007} have proposed the PG-means that produces clustering using the EM algorithm and uses the Kolmogorov-Smirnov test to decide $k$.
However, many methods to estimate $k$ cannot be applied to spherical data because they assume data generated from Gaussian distributions.

The purpose of the $k$-means is to minimize the sum of Euclidean distances between the data points and the centroids of the clusters to which the data points belong.
The methods using Euclidean distance can be considered to assume that data is sampled from a mixture of isotropic Gaussian distributions.
Because of the assumption, the $k$-means often make inadequate clustering of data generated from other distributions.
Many methods to find $k$ assume that data is generated from a mixture of Gaussian distributions.
Thus these methods may perform poor clustering of spherical data, which cannot be assumed to be generated from a Gaussian mixture.

The sk-means \cite{Dhillon:2001} can make proper clustering of spherical data.
It is one of the $k$-means family and uses cosine similarity instead of Euclidean similarity.
It assumes that data distribution is a mixture of von Mises-Fisher distributions.
It makes high-coherence clusters from text clustering \cite{Dhillon:2001}.
The expectation maximization algorithm for a mixture of von Mises-Fisher distributions is frequently used for geographical data \cite{Carta:2008,Masseran:2013}, trajectory data \cite{Calderara:2011}, and sound signal \cite{Ma:2015}.
These methods achieve clustering of spherical data, but they cannot find $k$.

\section{Methods}

\subsection{von Mises-Fisher Distribution}

The proposed method assumes that data points are generated from a mixture of von Mises-Fisher distributions.
The distribution is a typical probability density function for data on $d$-dimensional sphere.

A random vector $\bm x$ on a $d$-dimensional unit sphere ($\|\bm x\| = 1$) is said to be generated from the von Mises-Fisher (vMF) distribution
if its probability density function is denoted by
\begin{equation}
    \mathrm{vMF}(\bm x \mid \bm \mu, \kappa) = C_d(\kappa) \exp(\kappa \bm \mu^{\mathrm T} \bm x),
\end{equation}
where $\| \bm \mu \| = 1, \kappa \geq 0$.
$\bm \mu$ is the mean direction and $\kappa$ is the concentration parameter.
The normalizing parameter $C_d$ is given by
\begin{equation}
    C_d(\kappa) = \frac{\kappa^{d/2-1}}{(2 \pi)^{d/2}I_{d/2-1}(\kappa)},
\end{equation}
where $I_r$ is the modified Bessel function of the first kind and order $r$.

To estimate parameters of the von Mises-Fisher distribution, we can use maximum likelihood estimation.
The estimated $\bm \mu$ is obtained from
\begin{equation}
    \label{eq:mu}
    \bm \mu = \frac{\sum_i \bm x_i}{\| \sum_i \bm x_i \|}.
\end{equation}
The estimated ratio of the Bessel functions $A_d(\kappa)$ is obtained from
\begin{equation}
    A_d(\kappa)=\frac{I_{d/2}(\kappa)}{I_{d/2-1}(\kappa)} = \frac{\|\sum_{i=1}^N \bm x_i\|}{N} = \bar{R}.
\end{equation}
The detail of algebra to obtain these equations can be found in the papers \cite{Dhillon:2001,Banerjee:2005,Sra:2012}.

It is difficult to estimate the concentrate parameter $\kappa$ because an analytic solution cannot be obtain by maximizing likelihood estimation.
However, we can obtain the ratio of Bessel functions $A_d(\kappa)$ using maximum likelihood estimation.
To obtain approximation of $\kappa$ from the estimated ratio of Bessel functions $\bar{R}$, we use the simple equation proposed by Banerjee et al. \cite{Banerjee:2003} denoted by
\begin{equation}
    \label{eq:kappa}
    \kappa = \frac{\bar{R}(d-\bar{R}^2)}{1-\bar{R}^2}.
\end{equation}

\subsection{Spherical X-means (SX-means)}

The Spherical X-means (SX-means) is a method to find the number of clusters $k$.
The SX-means is based on the X-means.
It assumes that the data points in each cluster are sampled from a von Mises-Fisher distribution.
To decide $k$, the Bayes information criterion (BIC) is used.

The SX-means starts with a small number of clusters and grows the number $k$.
The estimation of $k$ is executed through two operations: the improve-parameters and the improve-structure operations.
The improve-parameters operation consists of two processes.
In the first process, each data point is assigned to a cluster using the spherical $k$-means (sk-means) which is a method for clustering spherical data.
The algorithm of the sk-means is shown in appendix A.
In the next process,  the BIC of each cluster is calculated assuming that data points in each cluster are sampled from a von Mises-Fisher distribution.
The BIC of the cluster $C_i$ is called preBIC$_i$.
The improve-structure operation consists of three processes.
In the first process, the data points in each cluster are assigned to the two subclusters using the sk-means.
In the next process, the BIC of each cluster is calculated assuming that data points in each cluster are sampled from two von Mises-Fisher distributions.
The calculated BIC of the cluster $C_i$ is called postBIC$_i$.
In the last process, if the postBIC$_i$ is more than the preBIC$_i$, the subclusters are accepted and $k$ is added 1.
Otherwise, the subclusters are rejected.
These operations are repeated until estimated $k$ in the last iteration equals to that in the current iteration.
The detailed algorithm of the SX-means is shown in Algorithm \ref{al:sx-means}.

\begin{algorithm}[tb]
\renewcommand{\algorithmicrequire}{\textbf{Input:}}
\renewcommand{\algorithmicensure}{\textbf{Output:}}
\caption{SX-means}
\begin{algorithmic}[1]
\REQUIRE Set $X$ of data points on $d$-dimensional sphere.
\ENSURE Estimated $k$ and clustering of $X$.
\STATE Initialize $k = 2$.
\REPEAT
    \STATE \{Improve parameter operation\}
    \STATE Run the sk-means.
    \FOR{$i = 1$ to $k$}
        %\STATE Estimate $\bm{\mu}_i$ and $\kappa_i$ using Eq. \ref{eq:mu} and \ref{eq:kappa}, respectively.
        \STATE Calculate preBIC$_i$.
    \ENDFOR
    \STATE \{Improve structure operation\}
    \STATE $k_\mathrm{new} = k$
    \FOR{$i = 1$ to $k$}
            \STATE Run the sk-means to assign the data points in $C_i$ to the subcluster, $c_1$ or $c_2$.
            \STATE Calculate postBIC$_i$.
            \IF{preBIC$_i$ $<$ postBIC$_i$}
                \STATE $k_\mathrm{new} \leftarrow k_\mathrm{new} + 1$
            \ENDIF
    \ENDFOR
    \STATE $k = k_\mathrm{new}$
\UNTIL Estimated $k$ in the current loop is equal to that in the last loop.
\STATE Run the sk-means.
\end{algorithmic}
\label{al:sx-means}
\end{algorithm}

\subsection{fixed SX-means}

More parameters make a model more complex and more frequently induce various problems such as the dead unit problem due to instability.
Fujita \cite{Fujita:2017} has achieved stable clustering of data in cylindrical coordinates by fixing the concentrate parameter $\kappa$.
In this study, to stabilize the proposed method, $\kappa$ is specified in advance.
In this study, the SX-means with the fixed-$\kappa$ model is called the fixed SX-means.
The algorithm of the fixed SX-means is the same as the SX-means except $\kappa$ is fixed.

\subsection{Bayesian Information Criterion}

The SX-means uses the Bayesian Information Criterion (BIC) to decide whether that data points assigned to a cluster are sampled from one von Mises-Fisher distributions or two those.
BIC is calculated from the following formula used in X-means \cite{Pelleg:2000}:
\begin{equation}
    \mathrm{BIC} = l - \frac{p}{2} \log R,
\end{equation}
where $l$ is the log likelihood function of the data points according to the model, $p$ is the number of parameters of the model, and $R$ is the number of data points.
This criterion is higher, and the model fits data points better.

In the improve-parameters operation, we can consider that the data points assigned to the $j$th cluster $X_j = \{\bm x_i \in C_j\}$ are sampled from one von Mises-Fisher distribution.
Given $\bm \mu_j$ and $\kappa_j$, we can write the probability density of the data points $X_j$ in the form:
\begin{equation}
    p(X_j \mid \bm \mu_j, \kappa_j) = \prod_{\bm x_i \in C_j} \mathrm{vMF}(\bm x_i \mid \bm \mu_j, \kappa_j).
\end{equation}
Viewed as a function of $\bm \mu_j$ and $\kappa_j$, the formula is the likelihood function for a von Mises-Fisher distribution.
The log-likelihood function of the $j$th cluster $C_j$ is denoted by
\begin{equation}
    \label{eq:likelihood_prebic}
    l_j = N_j \log C_d(\kappa_j) + \sum_{\bm x_i \in C_j} \kappa_j \bm \mu_j^T \bm x_i,
\end{equation}
where $N_j$ is the number of data points in $C_j$.
Thus, we can write the preBIC of $C_j$ as
\begin{equation}
    \mathrm{preBIC}_j = l_j - \frac{p}{2} \log N_j.
\end{equation}

In the improve structure operation, the data points in cluster $C_j$ are assumed to be generated from either two von Mises-Fisher distributions.
The data points in cluster $C_j$ are assigned to two subcluster $c_1, c_2$.
The point probability is denoted by
\begin{eqnarray}
P(\bm x_i \mid \bm \mu_m, \kappa_m) = \frac{n_m}{N_j} \mathrm{vMF}(\bm x_i \mid \bm \mu_m, \kappa_m),
\end{eqnarray}
where $n_m$ is the number of data points in the subcluster $c_m$.
The equation is derived from Pelleg and Moore \cite{Pelleg:2000}.
The log likelihood of $C_j$ is denoted by
\begin{eqnarray}
\label{eq:likelihood_postbic}
\nonumber l'_j &=& \sum_{m = 1}^2 \sum_{\bm x_i \in c_m} \log (\frac{n_m}{N_j} \mathrm{vMF}(\bm x_i \mid \bm \mu_m, \kappa_m)).
 \end{eqnarray}
Thus, the postBIC of $C_j$ is denoted by
\begin{equation}
    \mathrm{postBIC}_j = l'_j - \frac{p'}{2}\log N_j.
\end{equation}

It is difficult to calculate likelihood when the dimension of a data point is high because the numerator in the normalizing parameter $\kappa^{d/2 - 1}$ and the modified Bessel function in the denominator in the normalizing parameter rapidly grow with the dimension of a data point.
We cannot calculate the BIC because the rapid growth cause an overflow.
We, here, improve this problem by approximation of the modified Bessel function.
The approximation of the modified Bessel function is denoted by
\begin{equation}
    I_d(\kappa) \sim \frac{e^\kappa}{\sqrt{2\pi \kappa}}.
\end{equation}
This equation is the first term of Eq. 9.7.1 in \cite{Abramowitz:1970}.
Thus, approximation of the log likelihood functions Eq. \ref{eq:likelihood_prebic} and \ref{eq:likelihood_postbic} are respectively
\begin{equation}
    l_j = N_j (\frac{d-1}{2}(\log\kappa - \log(2\pi)) - \kappa) + \sum_{\bm x_i \in C_j} \kappa_j \bm \mu_j^T \bm x_i,
\end{equation}
and
\begin{eqnarray}
\nonumber    l'_j = \sum_{m=1}^2 \{n_m (\log n_m - \log N_j + \frac{d-1}{2}(\log\kappa_m - \log(2\pi)) - \kappa_m)\\
             + \sum_{\bm x_i \in c_m} \kappa_m \bm \mu_m^T \bm x_i\}.
\end{eqnarray}
These approximated likelihood does not have the terms growing exponentially.
We can calculate the log likelihood function and BIC even if dimension of a data point is very high.

\section{Results}

\begin{figure}[tb]
\centering
% For example, with the graphics package use
  \includegraphics[width=0.9\linewidth]{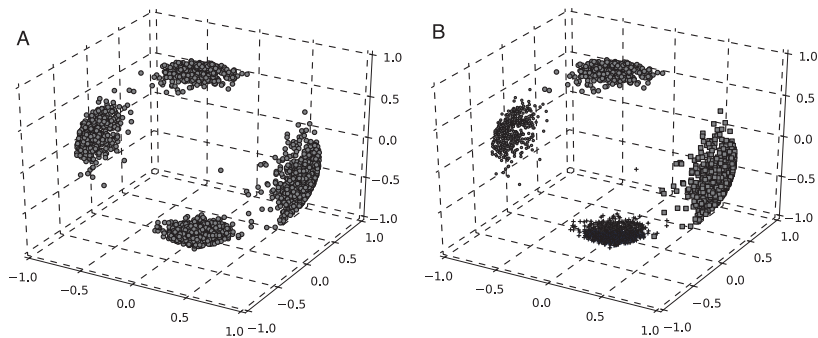}
\caption{Synthetic data and partitioned it.
(a) The data is generated from four von Mises-Fisher distributions which have parameters shown in Table \ref{tab:parameters}.
(b) Clustering result by the SX-means. The data points represented by the same symbol belong to the same cluster.}
\label{fig:clustering}       % Give a unique label
\end{figure}

The first experiment shows that the SX-means can estimate the parameters from synthetic data.
The synthetic data has 2000 data points from four von Mises-Fisher distributions on the three-dimensional sphere shown in Fig. \ref{fig:clustering}A.
The parameters of the input data are shown in Table \ref{tab:parameters}, where $i$ is the cluster number, $N_i$ is the number of data points in the cluster $i$, $\bm \mu_i$ is the mean vector of the cluster $i$, and $\kappa_i$ is the concentrate parameter of the cluster $i$.
Figure \ref{fig:clustering}B shows the clustering result of the SX-means.
This figure shows that the SX-means can find the number of clusters $k$ and make clustering of the data points.
Table \ref{tab:est para} shows the parameters estimated by the SX-means.
This result shows that the SX-means can precisely estimate the number of data points in each cluster, the centroid vectors, and $k$ from the synthetic data.
The concentrate parameters, however, could not be precisely estimated.
This reason is that estimation of $\kappa$ is more difficult than that of other parameters when the dimension of data points is low \cite{Sra:2012,Tanabe:2007}.

\begin{table}[tb]
\begin{minipage}{0.5\hsize}
\caption{Parameters of data set}
\begin{center}
\begin{tabular}{|c|c|c|c|}
\hline
$i$  & $N_i$ & $\bm \mu_i$ & $\kappa_i$\\
\hline
1 & 700 & $( 0, 0, -1)$ & 100\\
2 & 600 & $( 1, 0, 0)$ & 40\\
3 & 400 & $( -1, 0, 0)$ & 60\\
4 & 300 & $( 0, 0, 1)$ & 80\\
\hline
\end{tabular}
\end{center}
\label{tab:parameters}
\end{minipage}
\begin{minipage}{0.5\hsize}
\caption{Estimated parameters of data set}
\begin{center}
\begin{tabular}{|c|c|c|c|}
\hline
$i$  & $N_i$ & $\mu_i$ & $\kappa_i$\\
\hline
1 & 701 & $(0.004021, -0.001738, -1.0000)$ & 57.22\\
2 & 599 & $(0.9999, 0.004449, -0.01430)$ & 21.72\\
3 & 400 & $(-0.9999, 0.01388, -0.003918)$ & 34.29\\
4 & 300 & $(-0.006094, 0.01326, 0.9999)$ & 40.71\\
\hline
\end{tabular}
\end{center}
\label{tab:est para}
\end{minipage}
\end{table}

The second experiment shows $k$ estimated from three-dimensional synthetic data by the SX-means, the fixed SX-means with $\kappa = 10$ and $\kappa = 40$, and the X-means.
Each cluster in the synthetic data is generated from a von Mises-Fisher distribution with $\kappa = 100$ and has 500 data points.
The clusters' centroids are uniform randomly generated from points on three-dimensional sphere.
Table \ref{tab:est_k} shows $k$ estimated by the SX-means, the fixed SX-means, and the X-means.
Table \ref{tab:est_k} shows good agreement between the true $k$ and $k$ estimated by the SX-means.
$k$ estimated by the fixed SX-means with $\kappa = 10$ is smaller than the true $k$.
This smaller estimation will be evoked by the value of $\kappa$.
$\kappa$ of the fixed SX-means is smaller than that of the distribution generating the data, in other words, the fixed SX-means assumes broader distribution than that generating data.
Thus, data points in multiple distributions will be assigned to one cluster because of assuming broad distribution.
The fixed SX-means with $\kappa = 40$ shows the better performance than the fixed SX-means with $\kappa = 10$.
Especially, estimated $k$ for $k = 4$ and 6 is nearest to true $k$.
$k$ estimated by the fixed SX-means with $\kappa = 40$ is larger than it with $\kappa = 10$ for all cases.
The larger estimation is caused by the assumption of narrow distribution by the fixed SX-means with $\kappa = 40$.
$k$ estimated by the X-means is much larger than the true $k$.
For only X-means, the standard deviation of estimated $k$ is large.

\begin{table}[tb]
\caption{Estimated numbers of clusters and its standard deviations in parentheses}
\begin{center}
\begin{tabular}{|c|c|c|c|c|}
\hline
True $k$&        \multicolumn{4}{|c|}{Estimated $k$}       \\\hline
        &     SX-means         & fixed ($\kappa = 10$) & fixed ($\kappa = 40$) & X-means     \\\hline
2       & {\bf 2.000}  (0.000) & {\bf 2.000}  (0.000)  & {\bf 2.000}  (0.000)  & 3.200  (5.006)  \\\hline
3       & {\bf 3.050}  (0.497) & 2.650  (0.477)        & 2.850  (0.357)        & 3.750  (3.986)  \\\hline
4       & 3.700  (0.557)       & 3.450  (0.589)        & {\bf 3.750}  (0.622)  & 6.750  (6.730)  \\\hline
5       & {\bf 4.750}  (0.622) & 3.900  (0.624)        & 4.700  (0.458)        & 10.750  (8.887)  \\\hline
6       & {\bf 5.650}  (0.792) & 4.300  (1.005)        & {\bf 5.650}  (0.654)  & 11.750  (8.871)  \\\hline
7       & {\bf 6.400}  (0.583) & 4.950  (0.921)        & 6.100  (1.091)        & 18.500  (7.697)  \\\hline
8       & {\bf 7.200}  (1.030) & 5.150  (0.792)        & 7.100  (0.700)        & 21.000  (6.863)  \\\hline
9       & {\bf 8.300}  (1.100) & 5.800  (0.980)        & 7.850  (0.853)        & 19.100  (7.021)  \\\hline
10      & {\bf 8.800}  (1.568) & 5.750  (0.994)        & 8.700  (1.100)        & 22.600  (5.517)  \\\hline
11      & {\bf 10.300} (1.187) & 5.950  (1.117)        & 9.200  (0.980)        & 23.250  (3.740)  \\\hline
12      & {\bf 10.700} (1.005) & 6.350  (1.621)        & 10.300  (1.269)       & 24.850  (2.725)  \\\hline
\end{tabular}
\begin{tablenotes}
Estimated $k$ is the mean of twenty runs with randomly-generated synthetic data. The best estimations are bold. ``fixed'' in the table means the fixed SX-means.
\end{tablenotes}
\end{center}
\label{tab:est_k}
\end{table}

\begin{table}[tb]
\caption{Real-world Datasets}
\begin{center}
\begin{tabular}{|c|c|c|c|}
\hline
Dataset & $k$ & $n$  & $d$  \\ \hline
Blobs   &  3 & 1500  & 3 \\ \hline
Iris    &  3 & 150   & 4 \\ \hline
Wine    &  3 & 178   & 13 \\ \hline
Ecoil   &  8 & 336   & 7 \\ \hline
Yeast   & 10 & 1484  & 8 \\ \hline
MNIST   & 10 & 70000 & $28 \times 28$\\ \hline
CNAE-9  & 9  & 1080  & 856 \\ \hline
\end{tabular}
\end{center}
\label{tab:datasets}
\end{table}

The last experiment shows $k$ estimated from one synthetic dataset and six real-world datasets by the SX-means, the fixed SX-means with $\kappa = 20$ and $\kappa = 40$, and the X-means.
The synthetic dataset is Blobs generated using the function in scikit-learn, which is a machine learning library for Python.
The real-world datasets are Iris, Wine, Ecoil, Yeast, MNIST \cite{Lecun:1998}, and CNAE-9 dataset.
The characteristics of the datasets are exhibited in Table \ref{tab:datasets}.
The other parameters of Blobs are default values.
All datasets are zero-meaned.
All data points are normalized, but they are not normalized for only the X-means.
In this experiment, the normalize means that all data points are divided by the norms of themselves.

Table \ref{tab:est k of datasets} depicts $k$ estimated by the SX-means, the fixed SX-means with $\kappa = 10$ and $\kappa = 40$, and the X-means.
The SX-means shows the best performance for Wine and MNIST and the second-best for CNAE-9.
It, however, estimates much larger $k$ for Iris, Ecoil, and Yeast.
The fixed SX-means with $\kappa = 10$ presents the best performance for Blobs, Iris, and Yeast, and the second-best for Wine, Ecoil, and MNIST.
It, however, does not increase $k$ from the initial value for Wine and CNAE-9.
The fixed SX-means with $\kappa = 40$ estimates larger $k$ than with $\kappa = 10$.
This larger estimation is caused by the assumed distribution sharper than that for $\kappa = 10$.
The X-means shows the best performance for Ecoil and CNAE-9 and the second-best for Blobs and Iris.
However, $k$'s standard deviation of the X-means for CNAE-9 is the largest.
For Blobs, the standard deviation of $k$ estimated by the X-means is smallest.
This reason is that the distribution assumed by the X-means is consistent with Blobs.

\begin{table}[tb]
\caption{Estimated numbers of clusters from real-world datasets and its standard deviations in parentheses}
\begin{center}
\begin{tabular}{|c|c|c|c|c|c|}
\hline
dataset&True $k$ & SX-means            & fixed ($\kappa = 10$)   & fixed ($\kappa = 40$)& X-means \\\hline
Blobs  &   3     & 3.450 (1.857)       & {\bf 3.000} (0.949) &5.450 (3.398)  &2.950 (0.218) \\\hline
Iris   &   3     & 8.600 (1.715)       & {\bf 4.000} (0.000) &7.200 (0.748)  &7.100 (0.700) \\\hline
Wine   &   3     & {\bf 2.550} (0.497) & 2.000 (0.000)       &4.800 (0.600)  &22.200 (0.678) \\\hline
Ecoli  &   8     & 23.850 (2.762)      & 4.000 (0.000)       &21.900 (0.995) &{\bf 9.900} (1.179) \\\hline
Yeast  &   10    & 22.500 (1.857)      & {\bf 6.750} (0.433) &31.500 (0.742) &17.750 (4.763) \\\hline
MNIST  &   10    & {\bf 14.450} (1.071)& 5.400 (0.490)       &29.850 (0.477) &32.000 (0.000) \\\hline
CNAE-9 &   9     & 11.250 (2.426)      & 2.000 (0.000)       &2.600 (0.490)  &{\bf 6.800} (5.537) \\\hline
\end{tabular}
\begin{tablenotes}
Estimated $k$ is the mean of twenty runs with random initial values. The best estimations are bold. ``fixed'' in the table means the fixed SX-means.
\end{tablenotes}
\end{center}
\label{tab:est k of datasets}
\end{table}

\section{Conclusion} % (fold)
\label{sec:discussion}

This paper denotes a new method to estimate the number of clusters for data on $d$-dimensional sphere, called the SX-means.
For synthetic data generated from von Mises-Fisher distributions, the SX-means can precisely find the number of clusters.
The fixed SX-means also shows the moderate estimation.
For real-world data except Ecoil and CNAE-9, the fixed SX-means shows better performance than the X-means.
However, it is difficult to decide which methods, the SX-means or the fixed SX-means, is a better method.
The results of this study show that there are cases where the SX-means is better and where the fixed SX-means is better.

% section discussion (end)

\appendix
\section{sk-means (spherical $k$-means)}
\label{sec:vMF}

The sk-means has been developed for clustering of data on a spherical surface \cite{Dhillon:2001}.
The sk-means is the $k$-means adapted to spherical data.
It uses cosine similarity instead of Euclidean similarity.
Cosine similarity $S(\bm{\mu}_j, \bm{x}_i)$ between the centroid vector $\bm{\mu}_j$ and the data point $\bm{x}_i$ is calculated by
\begin{equation}
    S(\bm{\mu}_j, \bm{x}_i) = \bm{\mu}_j^{\mathrm T} \bm{x}_i,
\end{equation}
where  $\|\bm{\mu}_j\| = 1$ and $\|\bm{x}_i\| = 1$.
Let us consider the data set $X = \{\bm{x}_1, \bm{x}_2, ..., \bm{x}_N\}$.
The objective function of the sk-means $J$ is denoted by
\begin{equation}
    J = \sum_i^N \sum_j^K r_{ij} \bm{\mu}_j^{\mathrm T} \bm{x}_i.
\end{equation}
where $\bm{\mu}_j$ is the centroid vector of the cluster $C_j$ and $r_{ij} \in \{0, 1\}$ is binary indicator.
If $r_{ij} = 1$, the data point $\bm{x}_i$ is belong to the cluster $C_j$.
Otherwise, the data point $\bm{x}_i$ is not assigned to the cluster $C_j$.
The assignment of the data points to the clusters that maximizes the objective function is regarded as clustering result.
The sk-means maximizes the objective function through following steps:
\begin{enumerate}
    \item Specify $k$ and initialize $\bm{\mu}_j$.
    \item Assign each data point to the cluster with the nearest centroid.
    \item Calculate new centroid vectors of clusters. The centroid vector of the cluster $C_j$ is obtained by $\bm{\mu}_j = \sum_{x_i \in C_j} \bm{x}_i/\|\sum_{x_i \in C_j} \bm{x}_i\|$.
    \item Return step 2 if the assignment of data points change to clusters or the difference of the objective function between the current and the last loop is more than the threshold. Otherwise, finish the sk-means.
\end{enumerate}

%\bibliographystyle{plain2}      % mathematics and physical sciences
%\bibliography{clustering}

\end{document}